\documentclass[10pt,twocolumn,letterpaper]{article}

\usepackage{cvpr}
\usepackage{times}
\usepackage{epsfig}
\usepackage{graphicx}
\usepackage{amsmath}
\usepackage{amssymb}
\usepackage{subfig}

\usepackage{amsthm}
\usepackage{wrapfig}
\usepackage{algorithm}
\usepackage{algpseudocode}
\usepackage{array}
\usepackage{comment}

\usepackage[pagebackref=true,breaklinks=true,letterpaper=true,colorlinks,bookmarks=false]{hyperref}

\cvprfinalcopy 

\def\cvprPaperID{4478} 

\ifcvprfinal\fi
\begin{document}

\title{Conditional Single-view Shape Generation for Multi-view Stereo Reconstruction}

\author{
    Yi Wei$^{1,2}$\thanks{indicates equal contribution} \quad  Shaohui Liu$^{1,2*}$ \quad Wang Zhao$^{1,2*}$ \quad Jiwen Lu$^{1}$\thanks{corresponding author} \quad Jie Zhou$^{1}$ \\ 
	$^1$Department of Automation, Tsinghua University, Beijing, China \\
	$^2$Department of Electronic Engineering, Tsinghua University, Beijing, China \\
	{\tt\small b1ueber2y@gmail.com, \{wei-y15, zhaowang15\}@mails.tsinghua.edu.cn,} \\ {\tt\small \{lujiwen,jzhou\}@tsinghua.edu.cn}
}

\maketitle
\thispagestyle{empty}

\begin{abstract}
In this paper, we present a new perspective towards image-based shape generation. Most existing deep learning based shape reconstruction methods employ a single-view deterministic model which is sometimes insufficient to determine a single groundtruth shape because the back part is occluded. In this work, we first introduce a conditional generative network to model the uncertainty for single-view reconstruction. Then, we formulate the task of multi-view reconstruction as taking the intersection of the predicted shape spaces on each single image. We design new differentiable guidance including the front constraint, the diversity constraint, and the consistency loss to enable effective single-view conditional generation and multi-view synthesis. Experimental results and ablation studies show that our proposed approach outperforms state-of-the-art methods on 3D reconstruction test error and demonstrate its generalization ability on real world data. Code and data is available at \href{https://github.com/weiyithu/OptimizeMVS}{\color{cyan}{https://github.com/weiyithu/OptimizeMVS}}.
\end{abstract}
\vspace{-10pt}

\section{Introduction}

Developing generative models for image-based three-dimensional (3D) reconstruction has been a fundamental task in the community of computer vision and graphics. 3D generative models have various applications on robotics, human-computer interaction and autonomous driving, etc. Researchers have discovered effective pipelines on reconstructing scene structures \cite{hoiem2005automatic, saxena2009make3d} and object shapes \cite{Kar_2015_CVPR, Vicente_2014_CVPR}. Recently, inspired by the promising progress of deep learning on 2D image understanding and generation, much great work has been done on using differentiable structure to learn either volumetric or point cloud predictions from single-view \cite{Fan_2017_CVPR, Tulsiani_2018_CVPR, NIPS2016_6206} and multi-view \cite{choy20163d, NIPS2017_6640} images. 

Despite the rapid progress on the task of single-view image-based shape reconstruction, there remains a fundamental question: \emph{can a single image provide sufficient information for three-dimensional shape generation?} It is intuitive that in one picture took or rendered from a specific view, only the front of the object can be seen. Ideally, most existing methods implicitly assume that the reconstructed object has a relatively symmetrical structure, which enables reasonable guess on the back part. However, this assumption may not be true when it becomes much more complex in real world scenarios.

\begin{figure}
	\includegraphics[width=230pt]{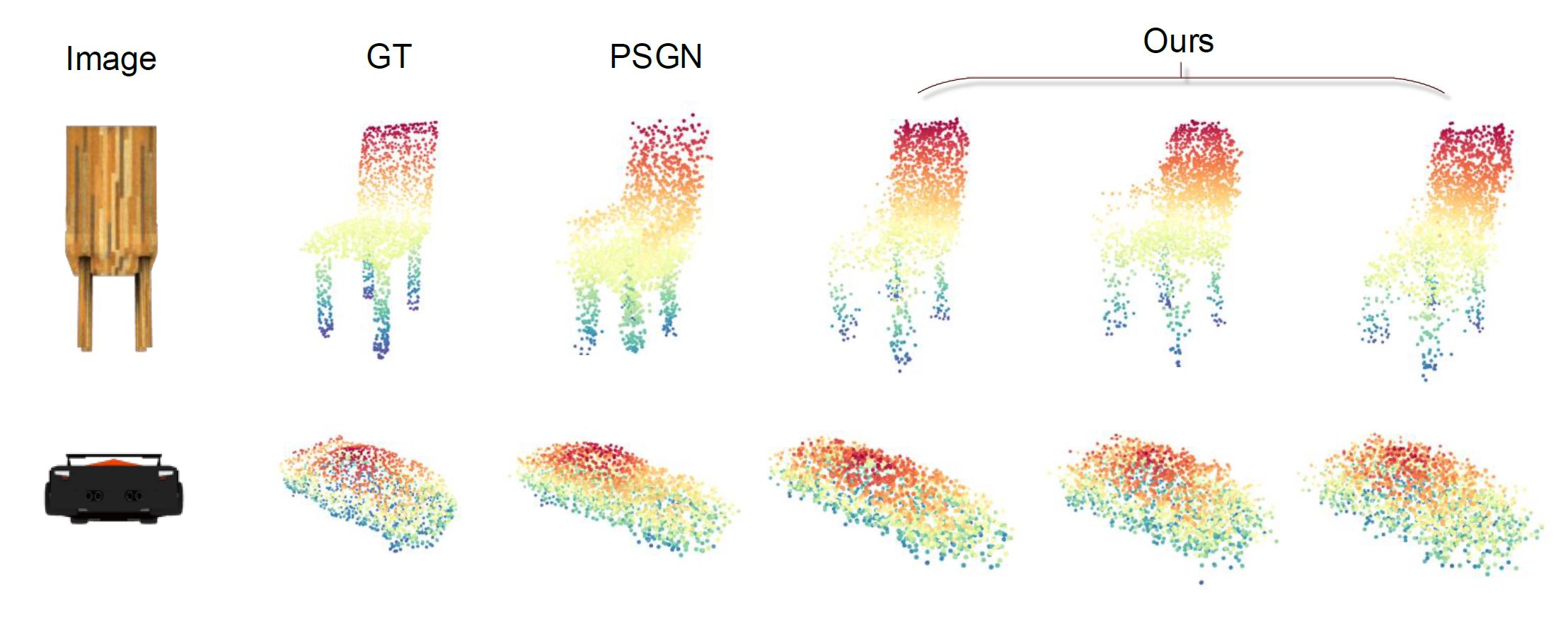}
	\vspace{-10pt}
	\caption{In real-world scenarios, one single image cannot sufficiently infer a single 3D shape due to occlusion. While our predictions can handle the ambiguity including the chair arm and the car length, deterministic methods can only predict one mean shape which is not necessarily correct. We further extend this for multi-view stereo reconstruction. }
	\label{fig::demo}
	\vspace{-15pt}
\end{figure}

In this paper, we address the problem of modeling the uncertainty for single-view object reconstruction. Unlike conventional generative methods which reconstruct shapes in a deterministic manner, we propose to learn a conditional generative model with a random input vector. As the groundtruth shape is only a single sample of the reasonable shape space for a single-view image, we use the groundtruth in a partially supervised manner, where we design a differentiable front constraint to guide the prediction of the generative model. In addition, we use a diversity constraint to get the conditional model to span the space more effectively. Conditioning on multiple random input vectors, our conditional model can give multiple plausible shape predictions from a single image. 

Furthermore, we propose a synthesis pipeline to transfer the single-view conditional model onto the task of multi-view shape generation. Different from most existing methods which utilize a recurrent unit to ensemble multi-view features, we consider multi-view reconstruction as taking the intersection of the predicted shape space on each single-view image. By introducing a simple paired distance metric to constrain the multi-view consistency, we perform online optimization with respect to the multiple input vectors in each individual conditional model. Finally, we concatenate the multi-view point cloud results to obtain the final predictions. 

Our training pipeline benefits from pre-rendered depth image and the camera pose without explicit 3D supervision. By modeling the uncertainty in single-view reconstruction via a partially supervised architecture, our model achieves state-of-the-art 3D reconstruction test error on ShapeNetCore \cite{shapenet2015} dataset. Detailed ablation studies are performed to show the effectiveness of our proposed pipeline. Additional experiments demonstrate that our generative approach has promising generalization ability on real world images.

\section{Related Work}
\paragraph{Conditional Generative Models:}

Generative models conditioned on additional inputs are drawing continuous attention with its large variety of applications. Conditional Generative Adversarial Networks (CGAN) \cite{mirza2014conditional} made use of the concept of adversarial learning and yielded promising results on various tasks including image-to-image translation \cite{Isola_2017_CVPR} and natural image descriptions \cite{Dai_2017_ICCV}. Another popular trend is the variational autoencoder (VAE) \cite{kingma2013auto}. Conditional VAEs achieved great success on dialog generation \cite{shen2017conditional, zhao2017learning}. Different from the CGAN \cite{mirza2014conditional} scenario, we only have limited groundtruth observations in the target space, which relates to the concept of one-shot learning \cite{lake2011one}. In this paper, we propose a partially supervised method with a diversity constraint to help learn the generative model. Then, we introduce a synthesis method on multiple conditional generative models.

\paragraph{Deep Single-view Reconstruction:}
With the recent advent of large 3D CAD model repositories \cite{shapenet2015,lim2013parsing,Sun_2018_CVPR,xiang_wacv14}, large efforts have been made on deep single-image reconstruction in 3D vision. While conventional methods \cite{choy20163d,Kar_2015_CVPR,NIPS2016_6096} focused on volumetric generation, point cloud and mesh representation were used in recent literature \cite{Fan_2017_CVPR,kurenkov2017deformnet,lin2017learning, groueix2018}. Researchers have introduced various single-view reconstruction approaches including 2.5D sketches \cite{NIPS2017_6657,genre,wu2018learning}, adversarial learning \cite{3D_Object_Rec,NIPS2016_6096}, generating novel views \cite{lin2017learning,Park_2017_CVPR,Shin_2018_CVPR,TDB16a}, re-projection consistency \cite{gwak2017weakly,Tulsiani_2018_CVPR,Tulsiani_2017_CVPR,NIPS2016_6206,Zhu_2017_ICCV}, high resolution generation \cite{johnston2017scaling,Tatarchenko_2017_ICCV} and structure prediction \cite{li2017grass,Niu_2018_CVPR}. Some recent post-processing attempts include point cloud upsampling \cite{Yu_2018_CVPR} and shape inpainting \cite{Wang_2017_ICCV}. These methods except \cite{Fan_2017_CVPR} all implicitly assumed that with prior knowledge the network can fantasy the missing part in the input image. However, in real world scenarios, the back part of the object may be much too complex to infer. This was recently addressed by \cite{wu2018learning}. In this work, we propose to model the ambiguity for the task of single-view point cloud reconstruction. Different from \cite{Fan_2017_CVPR} which used a relatively simple MoN loss to enable multiple predictions, we focus on different treatments between the front part and the back part of the object and improve its representation ability.

\paragraph{Deep Multi-view Synthesis:}
Multiple images took or rendered from different views contain pose-aware information towards 3D model understanding. Conventional method used multi-view cameras for 3D reconstruction via estimated correspondence \cite{triggs1999bundle,hartley2003multiple,mellado2014super,ranftl2018deep,kim2018recurrent}. For RGB-based multi-view reconstruction, recent deep methods \cite{choy20163d,NIPS2017_6640} mostly utilized a recurrent unit to integrate the extracted features from each single view. \cite{lin2017learning,TDB16a} used concatenation to get dense point cloud predictions. For a specific CAD model, the reconstruction results from different views should be consistent. This consistency was used \cite{gwak2017weakly,Tulsiani_2018_CVPR,Tulsiani_2017_CVPR,NIPS2016_6206} as a supervisory signal via re-projection for unsupervised single-view generative model training. Our method shares similar spirits with CodeSLAM \cite{bloesch2018codeslam}, which used joint optimization over the learned compact, optimizable representation. In this work, we introduce a multi-view synthesis techniques by online optimizing the multi-view consistency loss on point clouds in canonical coordinate space with respect to input codes of conditional models.

\begin{figure*}
	\centering
	\includegraphics[width=6.3in]{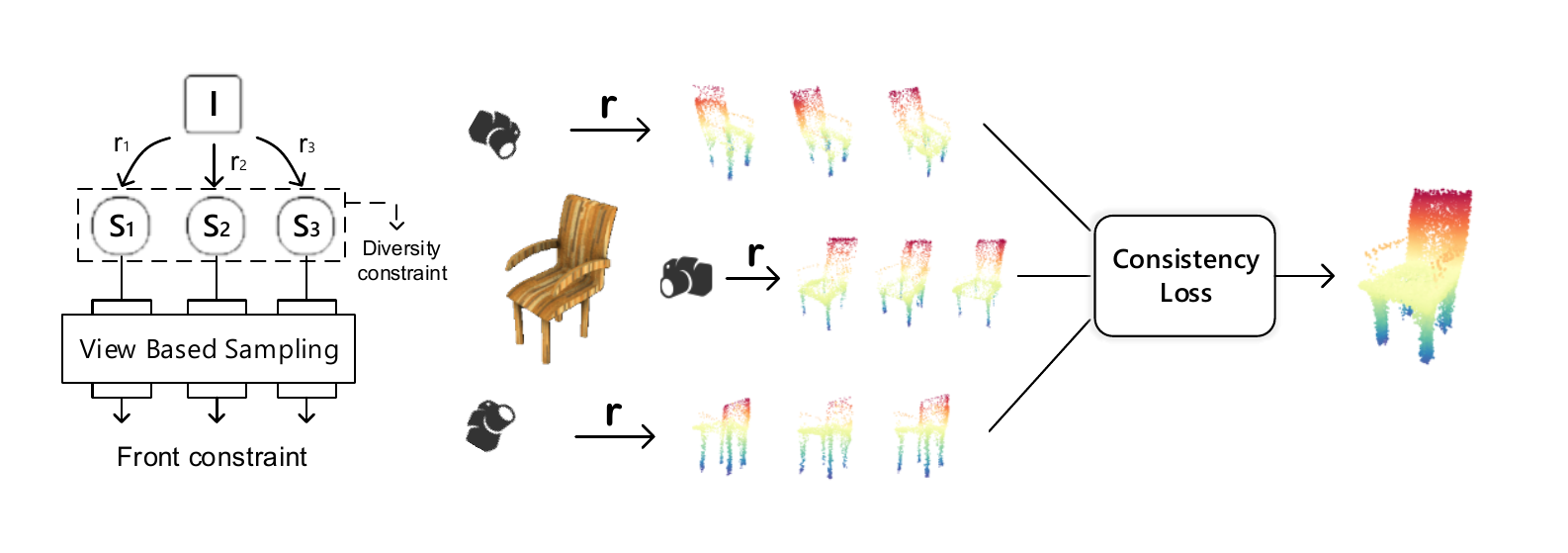}
	\vspace{-20pt}
	\caption{Overview of our proposed appoach. \textbf{Left:} the single-view training pipeline of the model. One single image is fed with a set of random inputs $\{r_i\}_{i=1}^n$ to get $S_i=f(I,r_i)$. Then, the partially supervised front constraint is used along with a diversity constraint to enable the model to focus more on the front part while maintaining generating diversity. \textbf{Right:} Inference. With different random inputs $\{r_i\}_{i=1}^n$, our conditional generative model $G: S=f(I,r|\theta)$ can generate multiple plausible shapes from each view. The consistency loss is used to synthesize the multiple conditional generative model to get the final predictions.}
	\label{fig::pipeline}
	\vspace{-15pt}
\end{figure*}

\section{Approach}
\subsection{Overview}
The problem of single-view shape reconstruction was conventionally formulated as a one-to-one mapping $\phi: I \rightarrow S$, where $I$ denotes the input RGB image and $S$ denotes the predicted shape. This one-to-one generative model was widely used to output either voxels \cite{choy20163d} or point clouds \cite{Fan_2017_CVPR} via cross entropy loss and differentiable distance metrics. Most existing methods took the implicit assumption that the input image is sufficient to predict the whole shape structure. 

Consider the probabilistic model $p(S|I)$, where $S$ is a random shape conditioned on the input image $I$. In perfect conditions where useful knowledge is completely learned from the groundtruth shape, the existing deterministic architecture $\phi: I \rightarrow S$ can learn the most probable shape $S^*$, where

\begin{equation}
\phi (I) = S^* = E_S[p(S|I)]
\end{equation}

Most existing single-view reconstruction methods utilized this deterministic formulation and could generalize relatively well to the test set. This is probably due to the fact that most objects in the widely used ShapeNet \cite{shapenet2015} dataset have a symmetric or category-specific structure which enables reasonable inference. However, this is arguably not true especially in complex scenarios. In fact, the structure of the occluded back part is usually relatively ambiguous. To better model this inherent ambiguity, we introduce a conditional generative model $f: I\times r \rightarrow S$, where the image-based generation is conditioned on a Gaussian input vector $r$. We aim to learn a mapping to approximate the probabilistic model $p(S|I)$ in the reasonable shape space. 

However, different from the scenarios of generation in CGAN \cite{mirza2014conditional}, we only have limited groundtruth (in fact, only one shape per image) which cannot span the reasonable shape space. Motivated by the fact that the front part can be sufficiently inferred from the input image while the back part is relatively ambiguous, we propose to train the conditional generative model in a partially supervised manner. Furthermore, we introduce a diversity constraint to help span the reasonable space. 

Let us take a step further into a multi-view scenario, where the problem comes to a many-to-one mapping. Considering only three input views for simplicity here, we have $I_1\times I_2\times I_3 \rightarrow S$. As most parts of the object are covered by different input views, we assume that multi-view reconstruction can be viewed as a deterministic inference with sufficient information. Given a conditional generative model $S=f(I,r)$, the output $S^*$ follow the constraint:

\begin{equation}
\label{eq::consis}
S^*=f(I_1,r_1)=f(I_2,r_2)=...=f(I_n,r_n)
\end{equation}

From the Bayesian perspective, Single-view and multi-view reconstruction can be formulated as to approximate $p(S|I_i)$ and $p(S|\{I_i\}_{i=1}^n)$. In this paper,
the idea is to first implicitly approximate $p(S|I_i)$ with a conditional model and then apply it to deterministic multi-view synthesis, which differs from conventional RNN-based methods \cite{choy20163d,NIPS2017_6640}. This choice has two main reasons. 1) The data scale is limited and only one 3D groundtruth exists for every image, making it not easy to explicitly parameterize $p(S|I_i)$. 2) For a 3D shape, rendered images from different views is correlated. Thus, it is relatively intractable to formulate $p(S|\{I_i\}_{i=1}^n)$ with $p(S|I_i)$ and directly optimize maximum likelihood. Same problem exists in many other research directions eg. multi-view pose estimation. We propose an alternate conceptual idea to get intersections of manifolds conditioned on different views with assistance of pair-wise distance minimization. Figure \ref{fig::pipeline} summarizes an overview of our proposed approach.

\subsection{Modeling the Uncertainty for Single-view Reconstruction}
\label{sec::uncertainty}

Given a specific architecture on a conditional generative model $S=f(I,r)$, we aim to learn the ambiguity of single-view reconstruction from limited groundtruth data. In this section, we first briefly review differentiable distance metric in the shape space, then introduce two of our proposed differentiable constraints to help learn the conditional model.

\paragraph{Distance Metric:}
Two existing differentiable distance metrics between point sets were originally used in \cite{Fan_2017_CVPR}. These metrics are Chamfer Distance (CD) and Earth Mover's distance (EMD) \cite{rubner2000earth}. CD finds the nearest neighbor and is formulated as below\footnote[3]{We use the first-order version of Chamfer Distance following \cite{lin2017learning}.} in Eq.(\ref{eq::chamfer}), while EMD learns a optimal transport between two point sets in Eq.(\ref{eq::emd}). 

\begin{equation}
\label{eq::chamfer}
d_{CD}(S_1, S_2)=\sum_{x \in S_1}{\min\limits_{y \in S_2} \Vert x-y \Vert_2}+\sum_{x \in S_2}{\min\limits_{y \in S_1} \Vert x-y \Vert_2}
\end{equation}

\begin{equation}
\label{eq::emd}
d_{EMD}(S_1, S_2)=\min\limits_{\phi: S_1 \rightarrow S_2} \sum_{x \in S_1} \Vert x-\phi (x) \Vert_2^2
\end{equation}

We use both of these two metrics in our training pipeline. Following \cite{lin2017learning}, the two terms in Chamfer Distance were jointly reported as pred$\rightarrow$GT and GT$\rightarrow$pred at test stage. 

\paragraph{Front Constraint:}
We propose a \emph{front constraint} along with a new differentiable operation: \emph{view based sampling}, which enables the conditional model to learn in a partially supervised manner. Different from recently proposed point cloud downsampling strategies \cite{Li_CVPR_2018, NIPS2017_7095} which aims to uncover inner relationship for coarse-to-fine understanding, our proposed \emph{view based sampling} layer outputs a set of points which consist the front part of the shape from a specific view.

\begin{figure}[tb]
	\includegraphics[width=8.2cm]{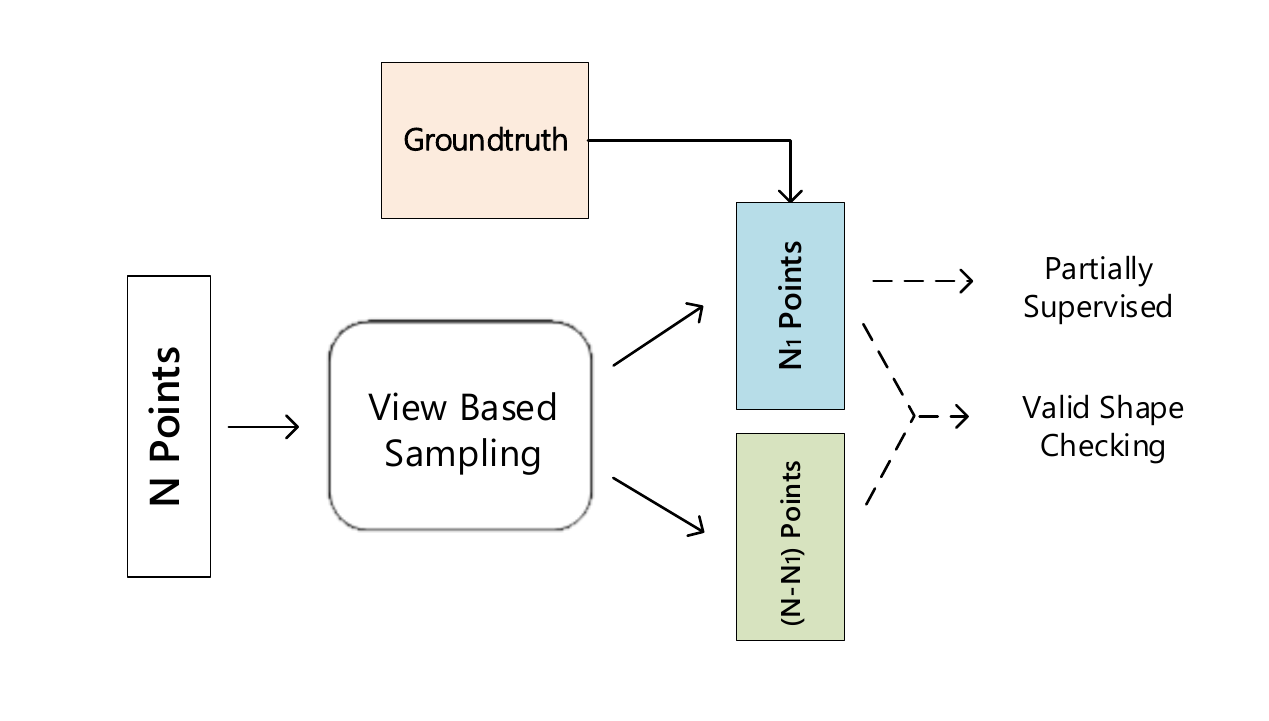}
	\vspace{-10pt}
	\caption{Partial supervision on the front part of the conditionally generated point clouds with \emph{view based sampling}. The validity (whether the outputs form a valid shape) is ensured by a GAN checking module in Eq.(\ref{eq::check_gan}).}
	\label{fig:sampling}
	\vspace{-10pt}
\end{figure}

The overview of the \emph{front constraint} is shown in Figure \ref{fig:sampling}. In the proposed approach, we get the generative model to focus more on the front $N_1$ points, while the $N-N_1$ remaining points are conditionally generated. For the \emph{view based sampling} layer, we first render the point cloud onto a 2D depth map with the intrinsic and extrinsic parameters given. Then, we sample all of the points which contribute to the rendered map. This strategy enables that all sampled points are on the front side of the object from the view. Note that because pixel-wise loss on the depth map used in \cite{lin2017learning} is only differentiable on the rendered $z$ axis, it will not work in our single-view training scenario (See Section \ref{sec::ablation}).

By either applying view based sampling to the groundtruth point cloud or applying inverse-projection to the pre-rendered depth map, we can get the groundtruth front part. Then, CD or EMD \cite{rubner2000earth} can be used to acquire the loss of the front constraint $loss_{front}$ and differentiably guide the sampled point cloud.

\paragraph{Diversity Constraint:}
Because for one input image, only one groundtruth shape is available at the training stage, simply training the conditional generative model with groundtruth constraints will hardly get the model to span the reasonable shape space. For different input vectors $r$, we aim to get different predictions which all satisfy the front constraint. With the hinge loss as in the widely used Triplet Loss \cite{schroff2015facenet} in face verification, we propose a \emph{diversity constraint} which uses the Euclidean distance of input $r$ as the distance margin in 3D space. 

Specifically, considering paired input vectors $r_1$, $r_2$ for a single training image $I$, we have two predicted point clouds $S_1=f(I, r_1)$ and $S_2=f(I, r_2)$. The loss of the \emph{diversity constraint} is formulated as in Eq.(\ref{eq::div}) below. 

\begin{equation}
loss_{div}=\max (0, \ \Vert r_1-r_2 \Vert_2 - \alpha EMD(S_1, S_2))
\label{eq::div}
\end{equation}

Because the counts of both point clouds are equal, we use EMD \cite{rubner2000earth} to measure the distance between $S_1$ and $S_2$. The hyper-parameter $\alpha$ helps control the diversity of the predicted point clouds.

\paragraph{Latent Space Discriminator:}
Combining the front constraint and the diversity constraint forms a initial paradigm for modeling the ambiguity of the single-view reconstruction. However, this paradigm puts little pressure on the back part. Thus, motivated by the generative adversarial networks \cite{goodfellow2014explaining} and recently proposed representation learning method \cite{achlioptas2017learning} on point clouds, we propose to add a latent space discriminator to better learn the shape priors. Specifically, we first train an auto-encoder on the point cloud domain. Then, we transfer the decoder to the end of our architecture and get it fixed. Finally, we apply WGAN-GP \cite{WGAN-GP} on the top of the latent space. Take $E_S$ as the encoder from the point cloud domain to the latent space, and $E_I$ as the encoder we use from the input image $I_i$ and the random noise $r_i$ to the latent variable $z_i$, the loss is formulated as below, where $S$ is the sampled point cloud from the dataset. 

\begin{align}
    \label{eq::check_gan}
    loss_{gan}&= -\mathbb{E}_{I_i \sim p_{data}, r_i\sim p(r)}[D(E_I(I_i, r_i))] \nonumber\\& +\mathbb{E}_{S \sim p_{data}}[D(E_S(S))] \\&- \lambda\mathbb{E}_{\hat{z} \sim p_{\hat{z}}}[({||\nabla _{\hat{z}}D(\hat{z})||}_2-1)^2]\nonumber
\end{align}

\paragraph{Training on Single-view Images:}
As discussed, we can train a conditional generative model using single input image with the optimization objective in Eq.(\ref{eq::loss}) at the training stage. $\beta$, $\gamma$ denotes the relative loss weight of the diversity loss and the GAN loss respectively.

\begin{equation}
\label{eq::loss}
loss=loss_{front}+\beta loss_{div} + \gamma loss_{gan}
\end{equation}
The training is performed in an iterative min-max manner as the widely-used GAN training strategy. The hyper-parameter $\alpha$ and $\beta$ modulates how far the generative model goes beyond the observed groundtruth. 

\subsection{Synthesizing Multi-view Predictions}
\label{sec::integration}

\paragraph{Finetuning on Multi-view Images:}
To get the network to learn more clues on the high level structure of the object, we finetune the single-view pretrained model on multi-view conditions. For synthesizing the multi-view point clouds at the training stage, we simply concatenate predicted point clouds from different views. Then, $loss_{front}$ was computed in different views on the concatenated results and $loss_{div}$ was computed in different random inputs to guide the training process. Specifically, for each shape, 8 views and 5 random inputs for each is used to train the model. Similar to the single-view training stage, we use the combined $loss$ in Eq.(\ref{eq::loss}) as our minimization objective.

\paragraph{Inference:}
As shown in Eq.(\ref{eq::consis}), from the deterministic perspective the multi-view reconstruction can be viewed as taking the intersection of the reasonable shape space conditioned on each input image. Thus, we propose a consistency constraint directly on the shape level. Consider a set of results $\{S_i\}_{i=1}^n$ from $n$ different views, where $S_i = f(I_i, r_i)$. The consistency loss is formulated in Eq.(\ref{eq::loss_consis}).

\begin{equation}
\label{eq::loss_consis}
loss_{consis}=\frac{2}{n(n-1)}\sum_{i=1}^{n-1}\sum_{j=i+1}^n CD(S_i, S_j)
\end{equation}

\begin{figure}[tb]
	\includegraphics[width=260pt]{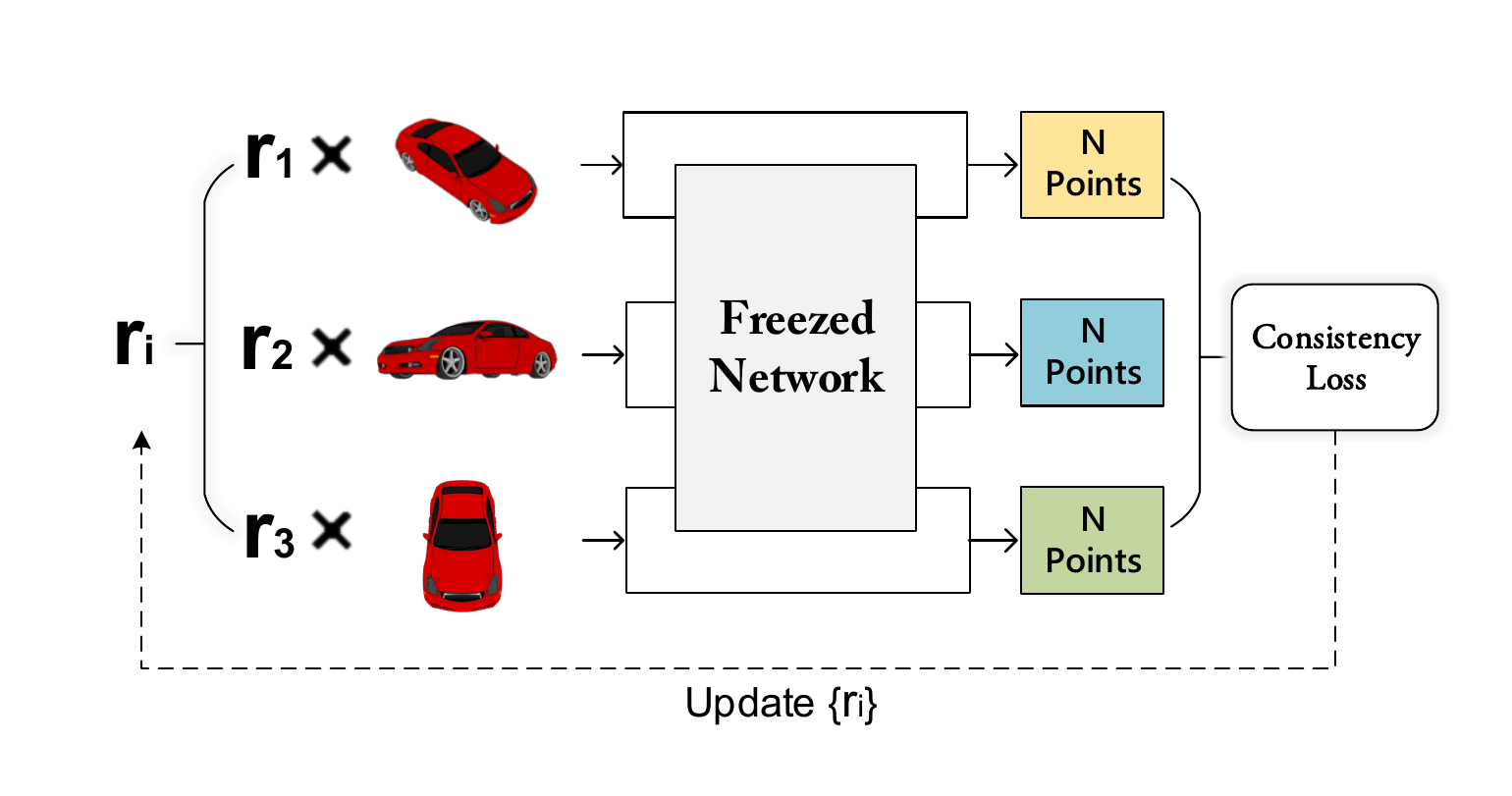}
	\caption{Multi-view inference by online minimizing the consistency loss.}
	\label{fig:fgsm}
	\vspace{-10pt}
\end{figure}

Figure \ref{fig:fgsm} shows our inference method. The method of freezing the inference model and adjusting the input is popular in the field of adversarial attacks \cite{goodfellow2014explaining}. By online minimizing $loss_{consis}$ with respect to the input vectors $\{r_i\}_{i=1}^n$ in the conditional model, we get more consistent results. To prevent the optimization from local minimum, we use heuristic search in the $\{r_i\}_{i=1}^n$ initialization. \textbf{Algorithm \ref{alg::inference}} shows our detailed inference pipeline. Our method does not require camera calibration at inference. 

\section{Experiments}
\subsection{Experimental Settings}
\begin{figure}[tb]
	\vspace{-10pt}
	\centering  
	\includegraphics[width=260pt]{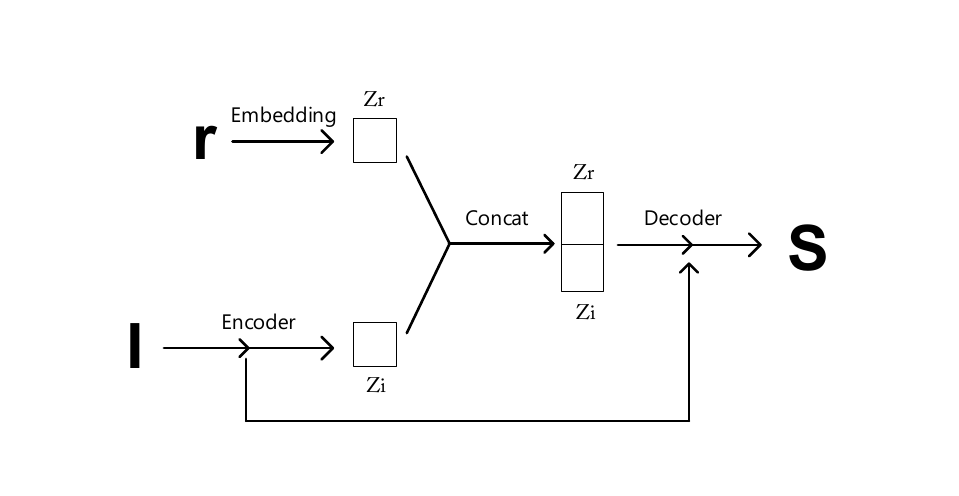}
	\vspace{-10pt}
	\caption{Brief overview of the network architecture.}
	\vspace{-10pt}
	\label{fig::arch}
\end{figure}

\paragraph{Network Architecture:}
Figure \ref{fig::arch} briefly shows our network architecture. For the encoder-decoder branch, we used the two-branch version of the point set generation network in \cite{Fan_2017_CVPR}. We set random input $r$ as a 128-dimensional vector. The embedding branch employs a structure with two fully connected layers and two convolutional layers. Channel-wise concatenation is performed on the embedded vector $z_r$ and the encoded features $z_i$. For more details on the two-branch network in \cite{Fan_2017_CVPR}, refer to our supplementary material.

\renewcommand{\algorithmicrequire}{\textbf{Input:}}
\renewcommand{\algorithmicensure}{\textbf{Output:}}
\begin{algorithm}[tb]
	\caption{Inference pipeline for multi-view reconstruction.}
	\label{alg::inference}  
	\begin{algorithmic}[1]
		\Require multi-view ($n$ views) images $\{I_i\}_{i=1}^n$, conditional generative model $G: S=f(I,r;\theta)$. 
		\Ensure predicted shape $S$. 
		\State Randomly sample 5 groups of $\{r_{ij}\}_{j=1}^5$, each of which consists of $n$ random inputs.
		\State Feedforward with $S_i=f(I_i, r_{ij};\theta)$ compute the $loss_{consis}$ in Eq.(\ref{eq::consis}) for each group. Denote the group with the minimum consistency loss $\{r^+_i\}_{i=1}^n$.
		\State Freeze the parameter $\theta$ of the inference model. Initialize $r_i=r^+_i$.  
		\State Iteratively minimize $loss_{consis}$ until convergence, get the optimized inputs $\{r^*_i\}_{i=1}^n$.
		\State Feedforward with $S_i=f(I_i, r^*_i;\theta)$ and concatenate $S_i$ to get the final prediction $S$.
	\end{algorithmic}
\end{algorithm}

\vspace{-8pt}
\paragraph{Implementation Details:}
We trained our conditional generative network for two stages on a GTX 1080 GPU. The input images were rendered from ShapeNetCore.v1 \cite{shapenet2015} with the toolkit provided by \cite{Tulsiani_2018_CVPR}. To cover the entire object, we uniformly sampled the rendered views along the horizontal circle with a random longitudinal perturbation. We took 80\% of the data for training and the rest for testing. We used $\lambda = 10$ and $\gamma = 0.1$ for adversarial learning. At the first training stage, we trained the model using single-view images for 40,000 iterations with a batch size 16 and 5 random inputs for each image. $\alpha_1 = 0.2$, $\beta_1 = 10.0$. Then, we finetuned our model for 100,000 iterations on multi-view images. There were 2 shapes in each batch, 8 views for each shape, and 5 random inputs for each view. $\alpha_2 = 0.1$, $\beta_2 = 1.0$. We used Adam with an initial learning rate 1e-4 in both stages. At test stage, we used 8 views to reconstruct the point clouds. The range of the longitudinal perturbation is a degree of $[-20,40]$. Following \cite{lin2017learning}, we scaled the reconstruction error CD by a factor of 100. Code is made available\footnote{https://github.com/weiyithu/OptimizeMVS}.

\begin{figure}[tb]
	\centering
	\vspace{-10pt}
	\includegraphics[width=240pt]{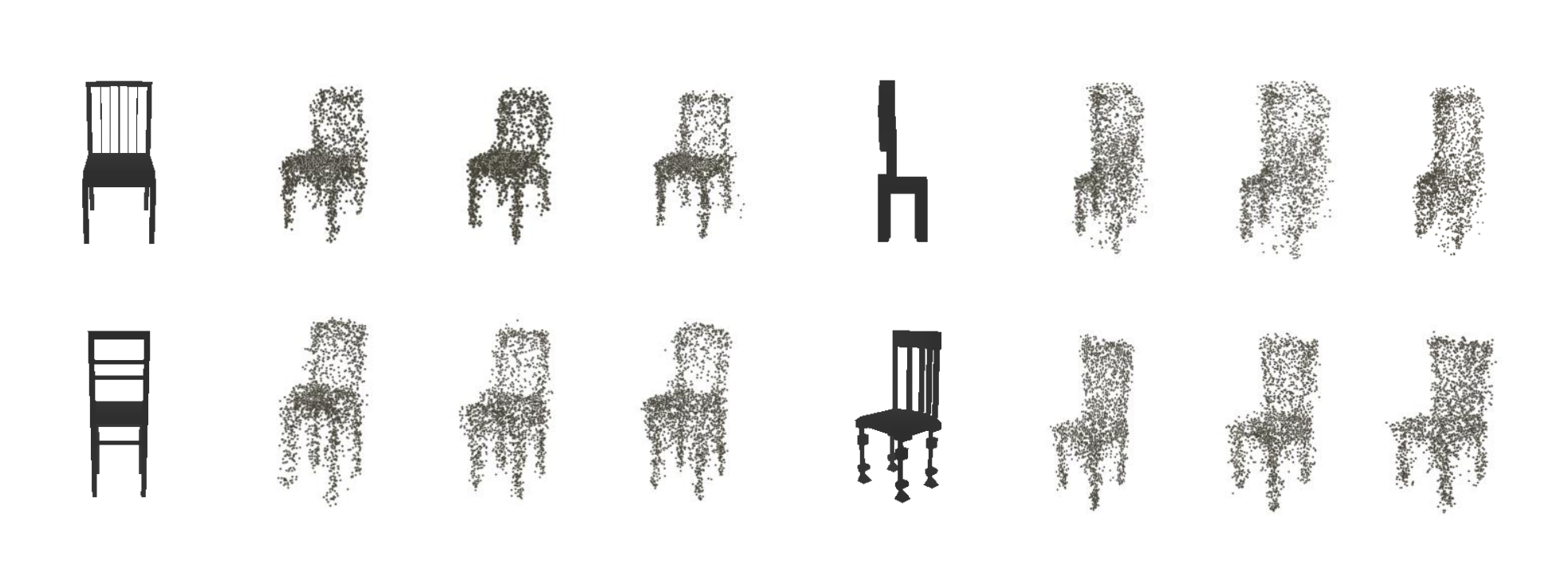}
	\caption{Visualization of multiple predictions on a single image conditioned on random sampled $r$.}
	\label{fig::vis_uncertainty}
	\vspace{-10pt}
\end{figure}

\begin{table}[tb]
	\centering
	\caption{Evaluation on the diversity of the conditional generative models. }
	\label{tab::uncertainty}    
	\begin{tabular}{l|c}
		\hline
		Method & Consistency loss \\
		\hline
		EMD + MoN \cite{Fan_2017_CVPR} & 1.65 \\
		\hline
		$loss_{front}$ & 0.55\\
		$loss_{front}$ + MoN & 2.52\\
		$loss$, $\beta=1.0$ & 2.88 \\
		$loss$, $\beta=5.0$ & 3.18 \\
		$loss$, $\beta=10.0$ & \textbf{3.36} \\
		\hline
	\end{tabular}
	\vspace{-10pt}
\end{table}

\subsection{Multiple Predictions on a Single Image}
Our generative model is able to predict multiple plausible shapes conditioned on the random input $r$. As discussed in Section \ref{sec::uncertainty}, the front constraint guides the generation of the front part, and the diversity constraint enables the conditional model to span the shape space. 

\paragraph{Qualitative Visualization:}
Figure \ref{fig::vis_uncertainty} visualizes the multiple predictions on a single input image conditioned on randomly sampled $r$. It is shown that our conditional model generates plausible shapes with a large diversity. The front part from the view of the input RGB image is predicted in a relatively more deterministic manner while the back part is mainly controlled by the random input $r$. 

\paragraph{Evaluation on Uncertainty Modeling:}
We conducted experiments to better verify the generating diversity of the proposed conditional generative model. We took one single image $I$ and randomly sampled 10 inputs $\{r_i\}_{i=1}^{10}$. Then, we fed the model with $S_i=f(I,r_i)$ and computed $loss_{consis}$ in Eq.(\ref{eq::consis}) on the predicted shape set $\{S_i\}_{i=1}^{10}$. We re-implemented the fully-supervised MoN method in \cite{Fan_2017_CVPR}. For fair comparison, we used the conditional model after the first training stage in this experiment. Table \ref{tab::uncertainty} shows the results. The partial supervision boosts the diversity of the predicted shapes. Moreover, it is demonstrated that when the loss weight $\beta$ of the diversity loss rises, the generating diversity gets consistent increase.

\begin{table}
	\centering
	\caption{CD (FPS-CD) results of single-category experiments on ShapeNet \cite{shapenet2015} dataset. We compare our methods with existing methods including \cite{choy20163d,Fan_2017_CVPR,lin2017learning,NIPS2016_6206}. }
	\label{tab::single-cat}
	\begin{tabular}{l||c|c|c}
		\hline
		Method & GT $\rightarrow$ pred & pred $\rightarrow$ GT & CD (FPS-CD) \\
		\hline
		3D-R2N2 & 2.47 & 3.21 & 5.68 \\
		PTN & 1.86 & 2.60 & 4.46 \\
		PSGN & 2.06 (2.06) & 2.27 (\textbf{2.27}) & 4.34 (4.34) \\
		Lin \emph{et al.} & 1.66 (2.16) & 2.35 (2.59) & 4.01 (4.75) \\
		\hline
		Ours & \textbf{1.39} (\textbf{1.73}) & \textbf{1.98} (2.35) & \textbf{3.37} (\textbf{4.08}) \\
		\hline
	\end{tabular}
\end{table}
\subsection{Multi-view Shape Reconstruction}
\begin{figure*}[tb]
	\centering
	\includegraphics[width=6.5in]{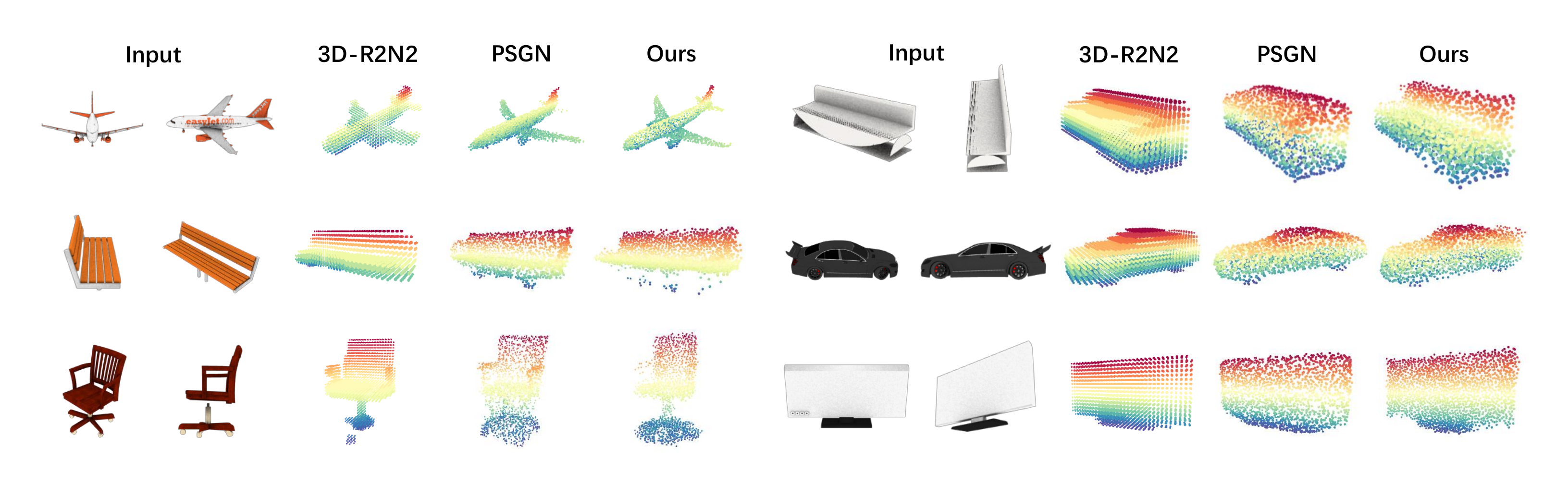}
	\caption{Qualitative comparison between ours and baseline approaches \cite{choy20163d,Fan_2017_CVPR}.}
	\label{fig::vis}
	\vspace{-10pt}
\end{figure*}

\paragraph{Evaluation Metric:}
The most widely used metric on evaluating point cloud generation is the Chamfer Distance in Eq.(\ref{eq::chamfer}). For comparison, we use the same protocol with \cite{lin2017learning}. However, it is worth noting that CD computation under different numbered point clouds is relatively confusing. Thus, we also report FPS-CD where we used farthest point sampling \cite{NIPS2017_7095} to get same-numbered point clouds.

\paragraph{Single-category Experiments:}
In this experiment, we applied our conditional generative model on the task of single-category multi-view shape reconstruction on ShapeNet \cite{shapenet2015} "chairs". We re-implemented several widely used image-based reconstruction methods including 3D-R2N2 (5 views) \cite{choy20163d}, PTN \cite{NIPS2016_6206}, PSGN \cite{Fan_2017_CVPR} and Lin \emph{et al.} \cite{lin2017learning} on our synthetic dataset. We converted the voxels predicted by \cite{choy20163d, NIPS2016_6206} to point clouds in the experiment. For the groundtruth point clouds, we used the uniformly sampled point clouds directly from \cite{achlioptas2017learning}. Note that our idea is also complementary to voxel-based deterministic methods (eg. MarrNet \cite{NIPS2017_6657}), where metrics can be developed on voxel space and back-propagation of cross-entropy loss is performed only from the front. Here we use PSGN \cite{Fan_2017_CVPR} with the point cloud outputs for direct comparison. Table \ref{tab::single-cat} shows the experimental results. It is reported that although our conditional generative method is not only partially supervised but also without explicit 3D supervision at training stage, our approach outperforms all baseline methods.
\begin{table}[tb]
	\centering
	\caption{CD (FPS-CD) results of multi-category experiments on ShapeNet \cite{shapenet2015} dataset. }
	\label{tab::multicat}    
	\begin{tabular}{c||c|c|c}
		\hline
		Category & 3D-R2N2 \cite{choy20163d} & PSGN \cite{Fan_2017_CVPR} & Ours \\
		\hline
		airplane & 5.25 & 2.89 (\textbf{2.89}) & \textbf{2.65} (3.10) \\
		\hline
		bench & 5.39 & 4.30 (4.30) & \textbf{3.48} (\textbf{4.17}) \\
		\hline
		cabinet & 4.60 & 4.87 (\textbf{4.87}) & \textbf{4.10} (5.39) \\
		\hline
		car & 4.51 & 3.68 (\textbf{3.68}) & \textbf{3.06} (4.03) \\
		\hline
		chair & 5.78 & 4.67 (4.67) & \textbf{3.80} (\textbf{4.64}) \\
		\hline
		display & 5.69 & 5.96 (5.96) & \textbf{4.44} (\textbf{5.27}) \\
		\hline
		lamp & 10.54 & 6.04 (\textbf{6.04}) & \textbf{5.15} (6.27) \\
		\hline
		loudspeaker & 6.54 & 6.42 (6.42) & \textbf{4.99} (\textbf{6.39}) \\
		\hline
		rifle & 4.38 & 3.22 (3.22) & \textbf{2.60} (\textbf{3.05}) \\
		\hline
		sofa & 5.43 & 4.93 (\textbf{4.93}) & \textbf{4.31} (5.35) \\
		\hline
		table & 5.31 & 4.45 (\textbf{4.45}) & \textbf{3.43} (4.51) \\
		\hline
		telephone & 5.06 & 4.34 (\textbf{4.34}) & \textbf{3.50} (4.35) \\
		\hline
		watercraft & 5.38 & 4.66 (4.66) & \textbf{3.57} (\textbf{4.24}) \\
		\hline
		all & 5.68 & 4.39 (4.39) & \textbf{3.58} (\textbf{4.34}) \\
		\hline
	\end{tabular}
\end{table}
\begin{table}[tb]
	\centering
	\caption{Comparison between the conditional model and the deterministic model. Both CD and FPS-CD are reported.}
	\label{tab::noinput} 
	\begin{tabular}{l||c|c}
		\hline
		Method & \ \ \ CD\ \ \  & \ \ FPS-CD\ \  \\
		\hline
		deterministic & 3.62 & 4.18 \\
		\hline
		conditional & \textbf{3.37} & \textbf{4.08} \\
		\hline
	\end{tabular}
	\vspace{-10pt}
\end{table}

\paragraph{Multi-category Experiments:}
We tested our model in multi-category experiments following \cite{choy20163d} on 13 popular categories on ShapeNet \cite{shapenet2015} dataset. As shown in Table \ref{tab::multicat}, our proposed method outperforms two baseline methods 3D-R2N2 \cite{choy20163d} and PSGN \cite{Fan_2017_CVPR} by a relatively large margin. 

\paragraph{Qualitative Results:}
For qualitative analysis, in Figure \ref{fig::vis} we visualize the predicted shapes for two state-of-the-art baseline methods: 3D-R2N2 \cite{choy20163d} and PSGN \cite{Fan_2017_CVPR}. It is shown that our partially supervised conditional generative model can infer reasonable shapes which are dense and accurate. More details are generated due to the specific aim on the front parts of the objects.

\subsection{Ablation Studies}
\label{sec::ablation}

\paragraph{Conditional vs. Deterministic:}
To demonstrate the effectiveness of the conditional model, we implemented a deterministic model $S=f_d(I)$. For fair comparison, we used an encoder-decoder structure similar with our network and trained the deterministic model for two stages with the front constraint. Single-category experiment was conducted on the deterministic model. Table \ref{tab::noinput} shows the results. Although the shape in ShapeNet \cite{shapenet2015} dataset often has symmetric structure, the conditional generative model outperforms the deterministic counterpart by 0.25 on CD. 

\begin{table}[tb]
	\centering
	\caption{Ablation studies on the diversity constraint and the consistency loss. “s1” denotes the pretraining on singleview images, while “s2” denotes the finetuning process on multi-view images. “s1” is always trained with diversity loss. The lossdiv in the table denotes the diversity loss specifically in “s2”. Experiments were conducted on the single-category setting. Both CD and FPS-CD is reported.}
	\label{tab::ablation}  
	\begin{tabular}{c|c|c|c||c|c}
		\hline
		s1 & s2 & $loss_{div}$ & $loss_{consis}$ & CD & FPS-CD \\
		\hline
		$\checkmark$ & $\times$ & $\times$ & $\times$ & 3.97 & 6.30 \\
		\hline
		$\checkmark$ & $\times$ & $\times$ & $\checkmark$ & 3.87 & 5.77 \\
		\hline
		$\checkmark$ & $\checkmark$ & $\times$ & $\times$ & 3.51 & 4.18 \\
		\hline
		$\checkmark$ & $\checkmark$ & $\checkmark$ & $\times$ & 3.40 & 4.16 \\
		\hline
		$\times$ & $\checkmark$ & $\checkmark$ & $\checkmark$ & 3.52 & 4.24 \\
		\hline
		$\checkmark$ & $\checkmark$ & $\checkmark$ & $\checkmark$ & \textbf{3.37} & \textbf{4.08} \\
		\hline
	\end{tabular}
	\vspace{-10pt}
\end{table}

\paragraph{Analysis on different features in the framework:}
We performed ablation analysis on three different features: two-stage training, diversity constraint at multi-view training stage and consistency loss during inference. As shown in Table \ref{tab::ablation}, all features achieve consistent gain on the final performance.

\paragraph{Front constraint vs. Projection loss:}
Our conditional model can be trained on single-view images with the front constraint and the diversity constraints. For comparison, we directly applied the projection loss used on multi-view  images training in \cite{lin2017learning} on single-view images, the training did not converge. Because the pixel-wise loss on the depth map suffers from non-differentiable quantization in the render process, the projection loss can only get gradients from the rendering axis. Our \emph{view based sampling} enables valid gradients on the x and y axis from the view. 

\begin{figure}[tb]
	\centering
	\vspace{-20pt}
	\includegraphics[width=0.45\textwidth]{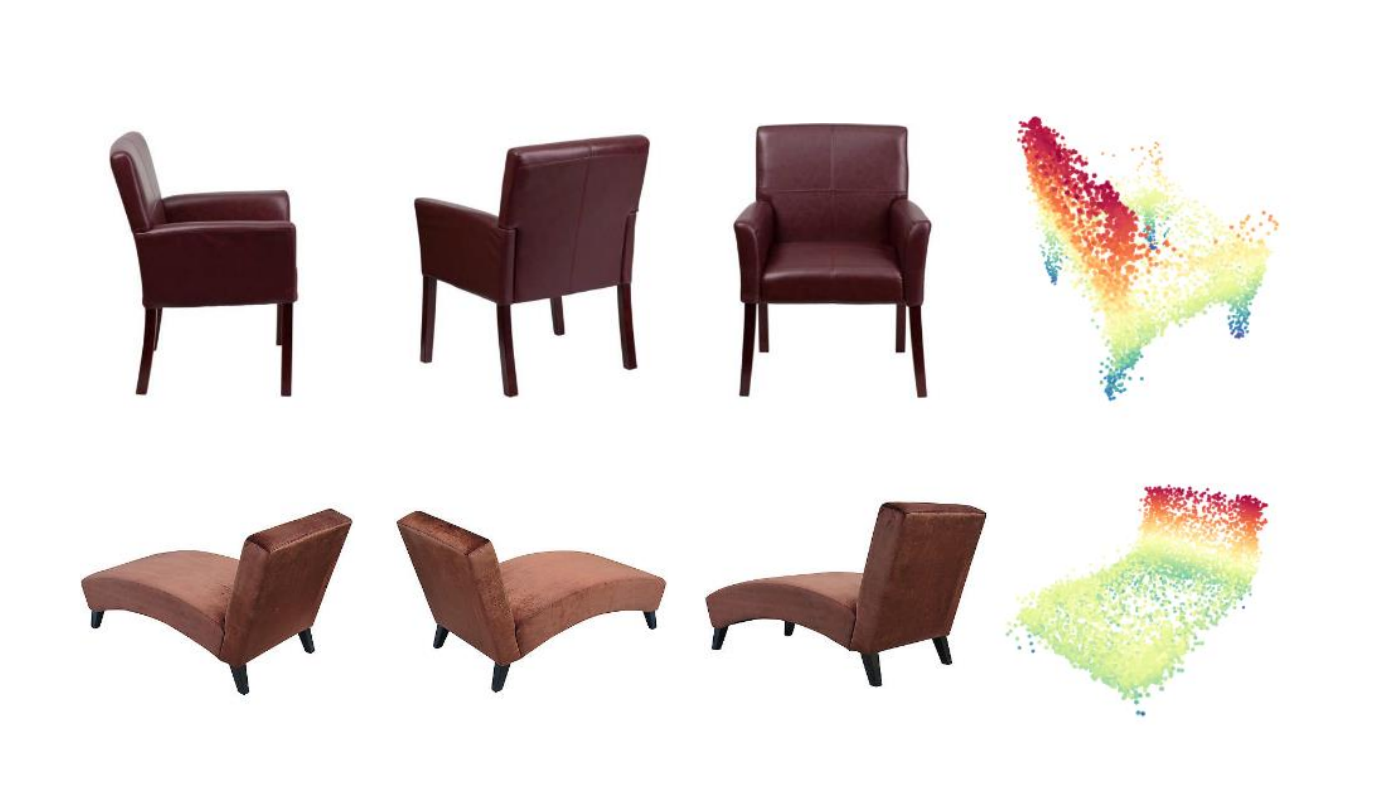}
	\vspace{-20pt}
	\caption{Visualization of the multi-view reconstruction results on real world images.}
	\label{fig::real}
\end{figure}

\begin{figure}[tb]
	\centering
	\vspace{-10pt}
	\includegraphics[width=200pt,height=140pt]{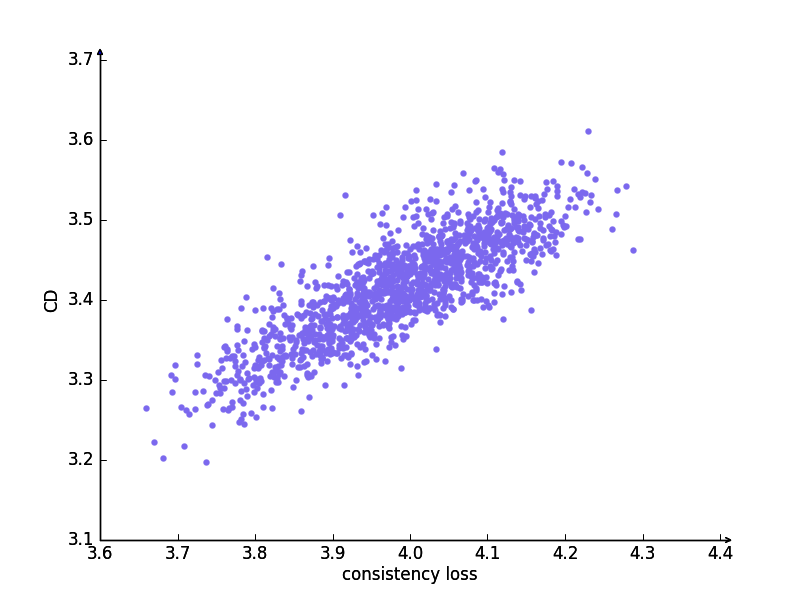}
	\caption{Correlation between the consistency loss and the 3D test error.}
	\vspace{-10pt}
	\label{fig::consis}
\end{figure}

\paragraph{Correlation between consistency loss and reconstruction error:}
In this part, we study the positive correlation between $loss_{consis}$ and the reconstruction error CD. First, we sampled $loss_{consis}$ and CD simultaneously at test stage. As shown in Figure \ref{fig::consis}, these two metrics show strong patterns of positive correlation. We further demonstrate this consistency on a highly diverse model (refer to our supplementary material for details). Table \ref{tab::consis} shows the experimental results. Minimizing the consistency loss gives consistent decrease on the CD metric with respect to the groundtruth shape. This demonstrates the fact that with $loss_{consis}$ as the mutual constraints inside the framework, the model will infer a more accurate shape at test stage. This verifies our interpretation on the success of applying the conditional generative model to the task of multi-view shape reconstruction.

\subsection{Reconstructing Real World Images}
The idea of multi-view reconstruction with our conditional generative model has great generalization ability. We conducted experiments on \emph{Stanford Online Products} dataset \cite{Song_2016_CVPR} for reconstructing real world images. Figure \ref{fig::real} visualizes our predictions. Our model generates surprisingly reasonable shapes by observing multi-view images in real world scenarios.

\begin{figure}[tb]
    \centering
    \includegraphics[width=0.5\textwidth]{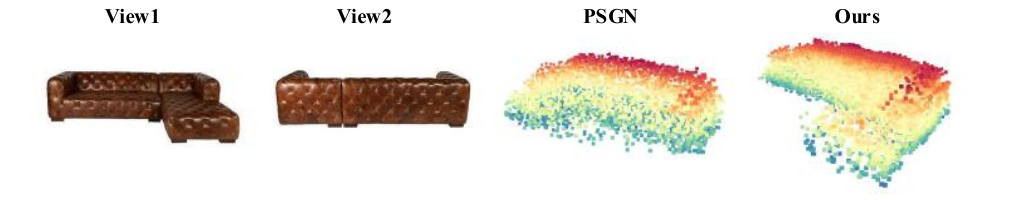}
    \caption{\textbf{Visualization on unsymmetric real-world data.} \textbf{Left Two:} Input images. For \cite{Fan_2017_CVPR}, we use both input images and then take the best prediction. For our model, we use n=2 input views. \textbf{Right Two:} While \cite{Fan_2017_CVPR} tends to hallucinate the back part symmetrically, our model achieves much better results, which further demonstrates the necessity of conditional modeling. }
    \label{fig:unsymmetric}
\end{figure}

To further demonstrate the necessity of conditional modeling, Figure \ref{fig:unsymmetric} shows visual results on unsymmetric real data sampled from \cite{Song_2016_CVPR}. While \cite{Fan_2017_CVPR} sticks to the symmetry prior and fails to generalize, our model generates a much realistic prediction.

\begin{table}[tb]
	\centering
	\caption{Study on the correlation between the consistency loss and the 3D test error CD. ``heuris" denotes the heuristic search in the $\{r_i\}_{i=1}^n$ initialization. ``bp" denotes the online optimization of $\{r_i\}_{i=1}^n$. ``dist1" denotes GT $\rightarrow$ pred and ``dist2" denotes pred $\rightarrow$ GT. Experiments on a specific model demonstrates the positive correlation between $loss_{consis}$ and CD. Both the heuristic search for initialization and the online update contribute to the performance improvement.}
	\label{tab::consis} 
	\begin{tabular}{c|c||c|c|c|c}
		\hline
		heuris & bp & dist1 & dist2 & CD & $loss_{consis}$ \\
		\hline
		no & no & 1.40 & 7.76 & 9.15 & 13.96 \\
		\hline
		yes & no & 1.41 & 3.24 & 4.65 & 5.96 \\
		\hline
		yes & yes & 1.40 & \textbf{2.80} & \textbf{4.21} & \textbf{4.66} \\
		\hline 
	\end{tabular}
\end{table}

\section{Conclusion}
In this paper, we have proposed a new perspective towards image-based shape generation, where we model single-view reconstruction with a partially supervised generative network conditioned on a random input. Furthermore, we present a multi-view synthesis method based on the conditional model. With the front constraint, diversity constraint and the consistency loss introduced, our method outperforms state-of-the-art approaches with interpretability. Experiments were conducted to demonstrate the effectiveness of our method. Future directions include studying the representation of the latent variables, rotation-invariant generation as well as better training strategies.

\section*{Acknowledgements}
This work was supported in part by the National Natural Science Foundation of China under Grant U1813218, Grant 61822603, Grant U1713214, Grant 61672306, and Grant 61572271. We sincerely thank Tianyu Zhao, Yu Zheng, Kai Zhong for valuable discussions.

{\small
\bibliographystyle{ieee}
\bibliography{pcgn}
}

\clearpage
\section*{Appendix}
\appendix
\begin{figure*}
	\centering
	\includegraphics[width=5.7in]{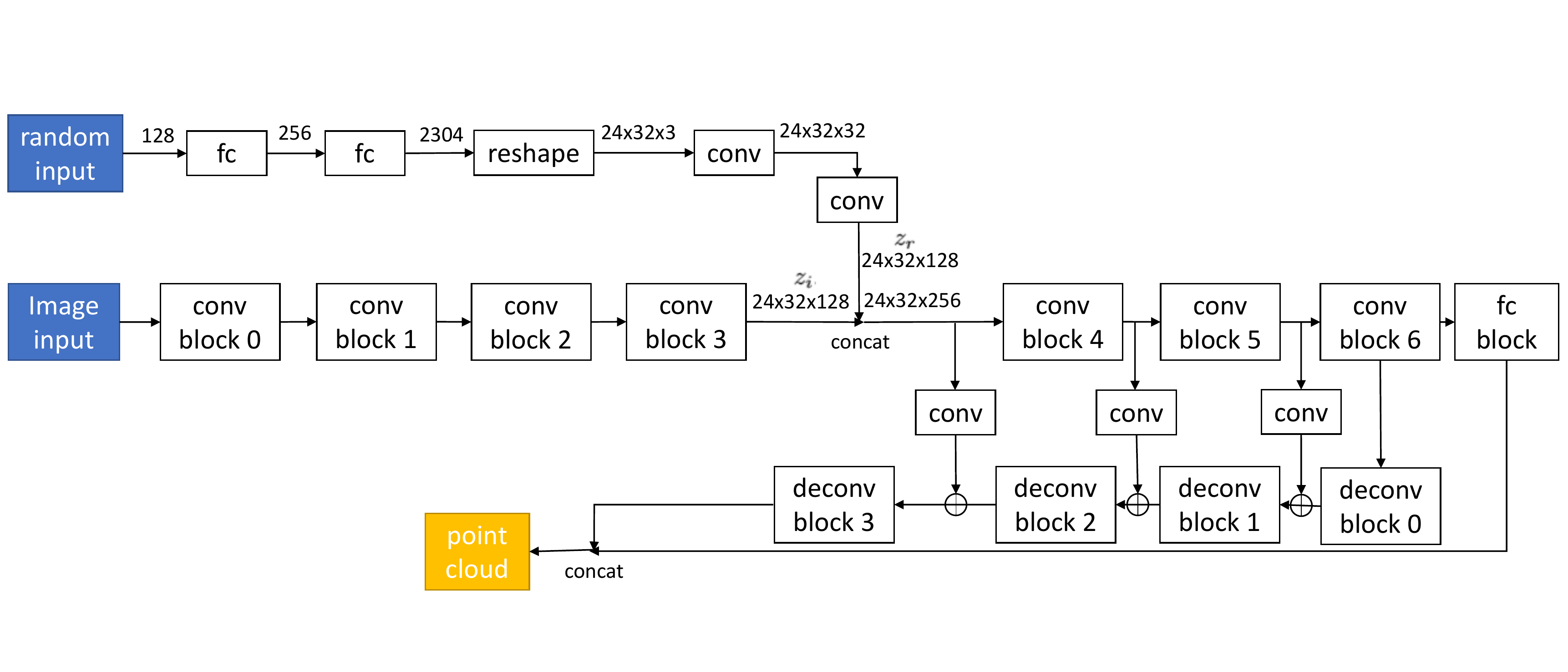}
	\caption{Network architecture of the conditional generative model.}
	\label{fig::arch_supp}
\end{figure*}

\section{Network architecture}
Our network architecture has two parts, one for embedding the conditional input $r$ to model uncertainty and the other for encoding and decoding input images. Figure \ref{fig::arch_supp} shows the encoder-decoder network architecture. For the embedding part, we set random input $r$ as a 128-dimensional vector. At the training stage and at the initialization of test stage, each dimension is sampled from a Gaussian distribution $N(0,1)$. Then we use two fully-connected layers (with 256 and 2304 output channels respectively), a reshape layer (from 2304 to 24$\times$32$\times$3) and two 3x3 convolutional layers (with 32 and 128 output channels respectively). For the encoder-decoder part, we use the two-branch version proposed in \cite{Fan_2017_CVPR}. This encoder-decoder network consists of two prediction branches, one is used for capturing high level structures and the other learns geometric continuity. 

First, the input image is encoded into an intermediate latent variable $z_i$ and the random input is embedded as $z_r$. Both $z_i$ and $z_r$ have the equal shapes of 24$\times$32$\times$128. Then, channel-wise concatenation is performed on the two encoded latent variables. Finally, the concatenated feature is decoded into the output point cloud.

\section{More Implementation Details}
\subsection{Datasets}
\paragraph{ShapeNet \cite{shapenet2015}} contains 57386 CAD models across 55 different categories. We randomly took 80\% of the objects for training and the rest for testing. For multi-view images rendering, we used the off-the-shelf renderer\footnote{\url{https://github.com/shubhtuls/mvcSnP/tree/master/preprocess/synthetic/rendering}} provided by \cite{Tulsiani_2018_CVPR}. For the groundtruth point clouds, we used the data\footnote{\url{https://github.com/optas/latent_3d_points}} provided by \cite{achlioptas2017learning}. Each point cloud consists of 2048 points uniformly sampled from the mesh on the dataset. We used "chair" for single-category experiments and the 13 popular categories following 3D-R2N2 \cite{choy20163d} for multi-category experiments. 

\paragraph{Stanford Online Products \cite{Song_2016_CVPR}}
is an online repository initially released to accelerate the field of metric learning. It contains automatically downloaded data from \url{https://www.ebay.com}. We used ``chair" and ``sofa" in our experiments for multi-view reconstruction on real world images. 

\subsection{Baseline approaches}
We reproduced several benchmark results of these methods on the datasets with their released code. In this section, we will show some details on these experiments.

\paragraph{3D-R2N2 \cite{choy20163d}}
For the 32$\times$32$\times$32 voxelized groundtruth for 3D-R2N2 \cite{choy20163d}, we directly used the provided voxels from their repositories. Following the paper, we applied two-stage training for 20k and 40k iterations on the training data. For Chamfer Distance computation, we uniformly sampled point clouds on the predicted voxels using their off-the-shelf functions.

\paragraph{PTN \cite{NIPS2016_6206}}
Similar to 3D-R2N2 \cite{choy20163d}, we uniformly sampled point clouds on the predicted voxels to enable comparsion with the groundtruth point clouds.

\paragraph{PSGN \cite{Fan_2017_CVPR}}
For fair comparison, we used the two-branch version of the network architecture described in \cite{Fan_2017_CVPR} with an output of 2048 points. We trained the fully-supervised deterministic model for 100k iterations with an Adam initial learning rate 1e-4. 

\paragraph{Lin \emph{et al.} \cite{lin2017learning}} We followed the two-stage training strategy in their paper. For the depth map rendered from the fixed 8 poses, we used their off-the-shelf released data described at \url{https://github.com/chenhsuanlin/3D-point-cloud-generation}. As they only released data for single-category experiments, we did not reproduce their 13-category results. Note that our input images and the groundtruth shapes are different from theirs (our groundtruth consist of 2048 points for each shape, which differs from their 10k dense point clouds). This made us unable to directly compare our 13-category performance with that reported in their main paper.
\begin{table}[htb]
	\centering
	\caption{Results of our model on different number of input views. `n' denotes the number of views. `cat1' and `cat13' denote single-category and multi-category experiments respectively. CD (FPS-CD) is reported.}
	\label{tab::number}    
	\begin{tabular}{p{1cm}<{\centering}||p{2cm}<{\centering}|p{2cm}<{\centering}}
		\hline
		n & cat13 & cat1 \\
		\hline
		1 & 5.76 (5.76) & 5.37 (5.37)\\
		\hline
		2 & 4.80 (4.93) & 4.57 (4.69) \\
		\hline
		3 & 4.32 (4.59) & 4.08 (4.33) \\
		\hline
		4 & 4.04 (4.44) & 3.79 (4.18) \\
		\hline
		5 & 3.92 (4.42) & 3.69 (4.18) \\
		\hline
		6 & 3.77 (4.37) & 3.54 (4.13) \\
		\hline
		7 & 3.67 (4.36) & 3.44 (4.12) \\
		\hline
		8 & 3.58 (4.34) & 3.37 (4.08) \\
		\hline
	\end{tabular}
\end{table}

\subsection{Highly diverse generative model design}
We present details on the highly diverse model used in the final paragraph of Section 4.4 in our main paper. To better demonstrate the positive correlation between the consistency loss and the 3D reconstruction error, we trained a highly diverse conditional generative model on multi-view images. Specifically, we applied diversity constraint on the whole concatenated point clouds at the second stage of the training. We used $\alpha=15.0$ and $\beta=0.5$ in this experiment. Similar to the main experiment, we trained the model for 40,000 iterations using Adam with an initial learning rate 1e-4.

From the Table 6 in the main paper we can infer the positive correlation between the consistency loss and CD at inference stage. Moreover, it is shown that applying the diversity constraint to the single-view predicted point clouds rather than the concatenated results gives much higher performance (3.37 vs. 4.66 for CD).

\section{More Ablation Studies}

\subsection{Ablation on number of input images}
We conducted experiments with different number of input views. We randomly sample $n$ views and run inference on both single and multiple categories. Results are shown in Table \ref{tab::number}. When we input only one view , the consistency loss is unable to work, so the performance of the conditional model is relatively poor. With more views observed, the performance becomes consistently better.

\subsection{Runtime analysis}
We did not use any type of connectivity on the view-based sampling layer. On 2048 (10k) points, our layer, which takes 6.6ms (10.2ms) on average, is an $O(n)$ approximation with $10.4\%$ ($3.6\%$) hidden parts included. The accurate mesh-based sampling is at least $O(nlgn)$ with a large constant. Generating a triangle mesh from 2048 (10k) points already takes 209.8ms (1.12s), which becomes the speed bottleneck at both training and inference. 

Note that in some cases the current system suffers from the problem of empty faces. It could be due to that the current view-based sampling is an approximation form which might wrongly samples the points from the back part. This will get the wrongly sampled points to be closer to the front, resulting in unbalanced density. Developing efficient usage of connectivity to better approximate the view-based sampling process might help resolve this issue.

\section{More Qualitative Results}
In this section, we show more samples of qualitative results on single-view conditional predictions and multi-view reconstruction.

\clearpage
\begin{figure*}[tb]
	\centering
	\includegraphics[width=5.7in]{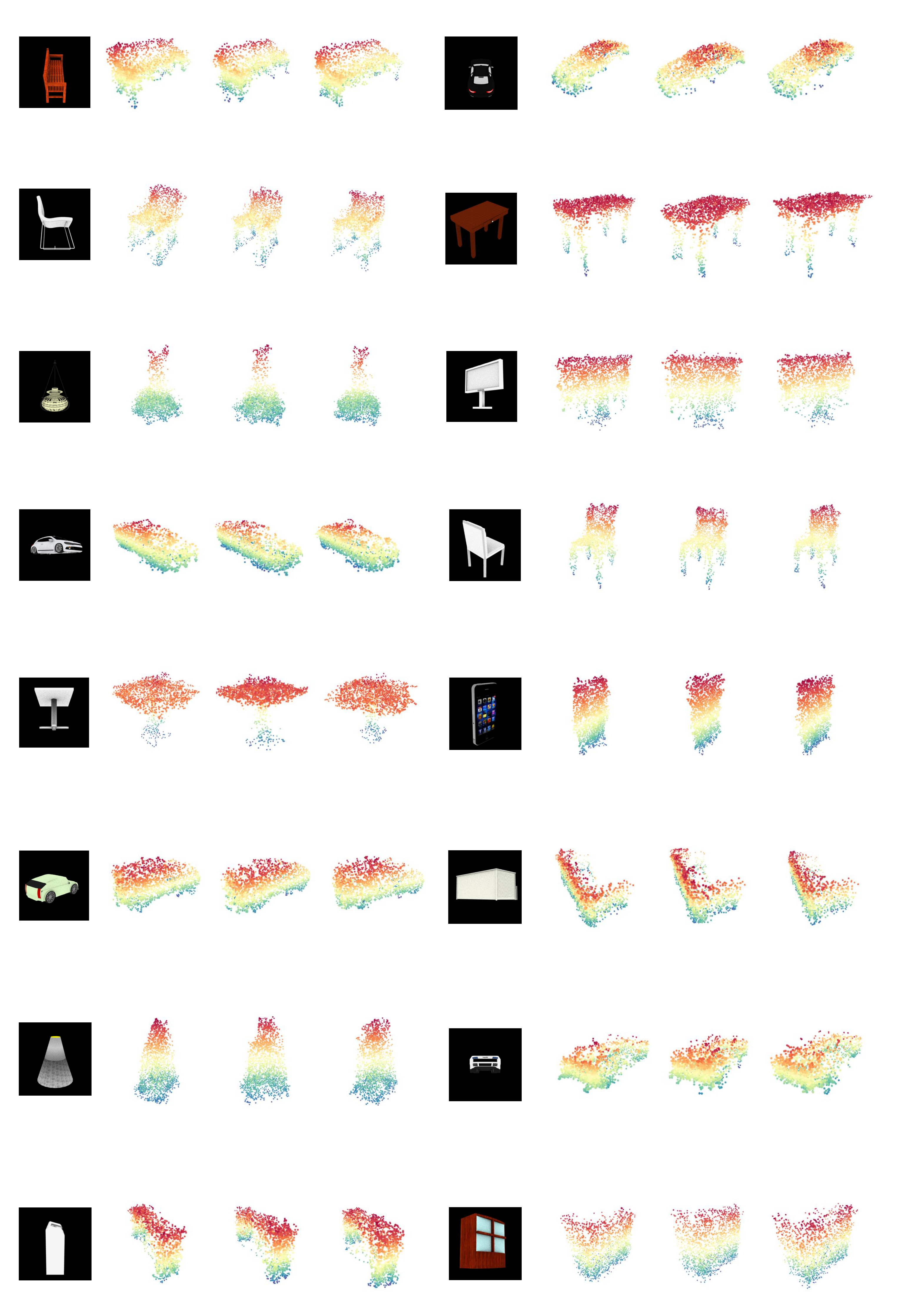}
	\caption{Visualization on multiple predictions on single-view images.}
	\label{fig::vis_supp_1}
\end{figure*}

\clearpage
\begin{figure*}[tb]
	\centering
	\includegraphics[width=5.7in]{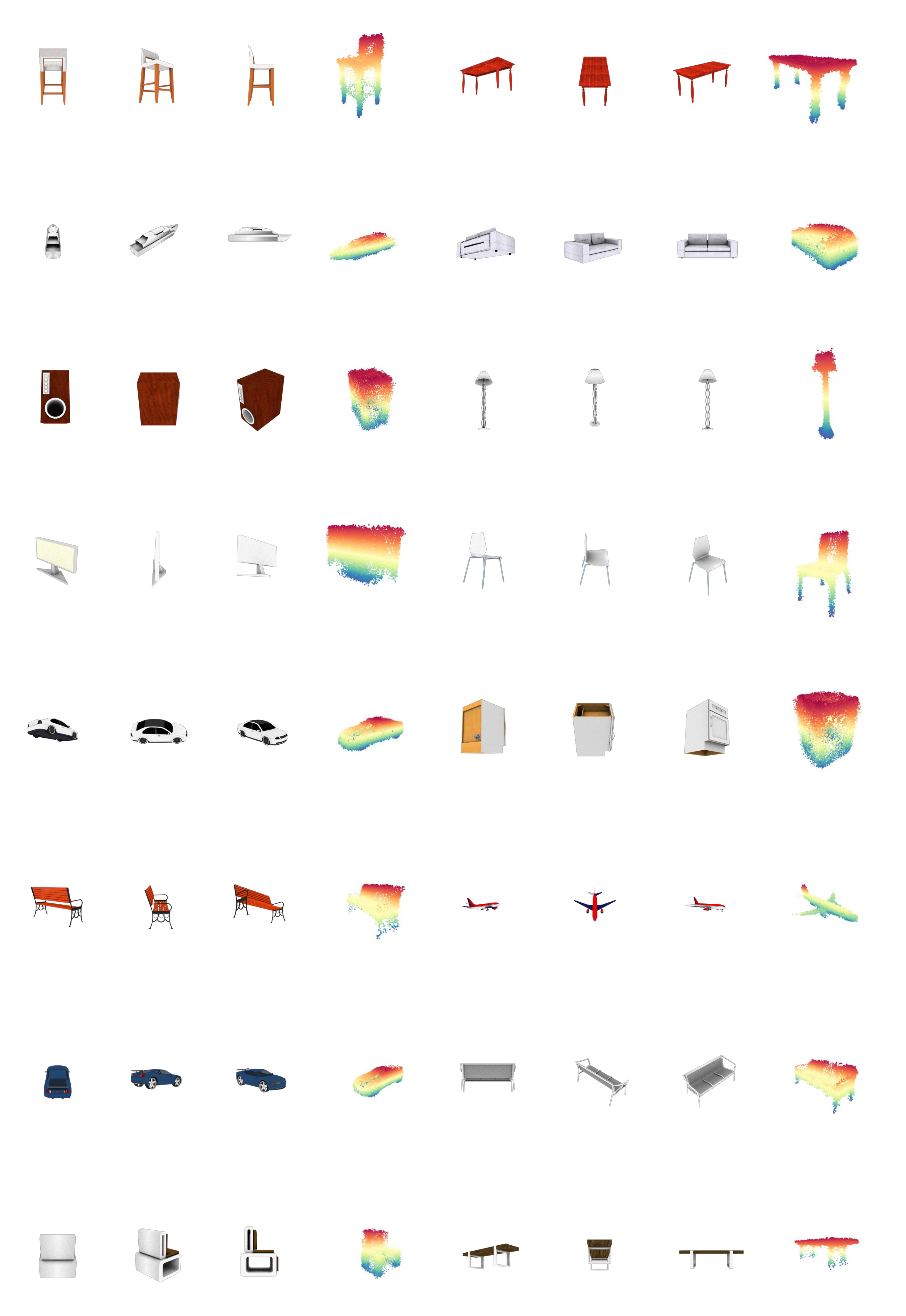}
	\caption{Visualization on multi-view reconstruction with our proposed method.}
	\label{fig::vis_supp_2}
\end{figure*}

\end{document}